\documentclass[10pt,twocolumn,letterpaper]{article}

\usepackage{iccv}
\usepackage{times}
\usepackage{epsfig}
\usepackage{graphicx}
\usepackage{amsmath}
\usepackage{amssymb}
\usepackage{float}
\usepackage{stfloats}
\usepackage{dsfont}

\usepackage[pagebackref=true,breaklinks=true,letterpaper=true,colorlinks,bookmarks=false]{hyperref}

\DeclareMathOperator*{\argmin}{arg\,min}

\iccvfinalcopy 

\def\iccvPaperID{27} 

\ificcvfinal\fi

\begin{document}

\title{Group-Conditional Conformal Prediction via Quantile Regression Calibration for Crop and Weed Classification}

\author{Paul Melki\\
IMS, CNRS, University of Bordeaux\\
EXXACT Robotics \\
{\tt\small paul.melki@u-bordeaux.fr}
\and
Lionel Bombrun\\
IMS, CNRS, University of Bordeaux\\
Bordeaux Sciences Agro\\
{\tt\small lionel.bombrun@ims-bordeaux.fr}\\
\and
Boubacar Diallo, Jérôme Dias\\
EXXACT Robotics \\
{\tt\small boubacar.diallo@exxact-robotics.com}\\
{\tt\small jerome.dias@exxact-robotics.com}
\and
Jean-Pierre Da Costa\\
IMS, CNRS, University of Bordeaux\\
Bordeaux Sciences Agro\\
{\tt\small jean-pierre.dacosta@ims-bordeaux.fr}\\
}

\maketitle
\ificcvfinal\thispagestyle{empty}\fi

\begin{abstract}
   As deep learning predictive models become an integral part of a large spectrum of precision agricultural systems, a barrier to the adoption of such automated solutions is the lack of user trust in these highly complex, opaque and uncertain models. Indeed, deep neural networks are not equipped with any explicit guarantees that can be used to certify the system’s performance, especially in highly varying uncontrolled environments such as the ones typically faced in computer vision for agriculture. 
   
   Fortunately, certain methods developed in other communities can prove to be important for agricultural applications. This article presents the conformal prediction framework that provides valid statistical guarantees on the predictive performance of any black box prediction machine, with almost no assumptions, applied to the problem of deep visual classification of weeds and crops in real-world conditions. The framework is exposed with a focus on its practical aspects and special attention accorded to the Adaptive Prediction Sets (APS) approach that delivers marginal guarantees on the model's coverage. Marginal results are then shown to be insufficient to guarantee performance on all groups of individuals in the population as characterized by their environmental and pedo-climatic auxiliary data gathered during image acquisition. 

   To tackle this shortcoming, group-conditional conformal approaches are presented: the ``classical” method that consists of iteratively applying the APS procedure on all groups, and a proposed elegant reformulation and implementation of the procedure using quantile regression on group membership indicators. Empirical results showing the validity of the proposed approach are presented and compared to the marginal APS then discussed.
\end{abstract}

\section{Introduction}
Artificial intelligence has become an integral component of precision agriculture systems. It provides the ``analytical" machinery that has allowed precision agriculture to adapt to the ever-increasing flow of data characterized by a high diversity of modalities (such as RGB images, LiDAR, text and GNSS) from multiple sources influenced by a large spectrum of natural and technical conditions. From this growing pool of raw data, machine learning algorithms, and particularly deep neural networks, have proven themselves to be the approach \textit{par excellence} to extract useful information. This information will either be directly turned into useful insight and decisions by human actors, or will flow through fully-automated robotic pipelines in autonomous agricultural systems \cite{zhang_agricultural_2021}.

Complex machine learning models have replaced classical ``handcrafted" models that were characterized by their well-defined interpretable features and their direct inspiration from agricultural and bio-environmental factors. Both practitioners and scientists in precision agriculture were comfortable with these classical approaches \cite{saia_transitioning_2020}. Deep learning models, on the other hand, with their complex components and architectures are not only relatively opaque to the agricultural community, but also, to a certain extent, to their own designers and developers \cite{zhang_survey_2021, szegedy_intriguing_2014}. While their performance prowess has been and still is being proven in the lab and in the field, some important issues such as interpretability \cite{zhang_survey_2021, selvaraju_grad-cam_2020, condran_machine_2022}, generalization to new observations and domains \cite{zhang_understanding_2017, zhang_understanding_2021, you_universal_2019}, robustness to noise and out-of-distribution observations \cite{madry_towards_2018, raghunathan_understanding_2020, arjovsky_out_2019}, and uncertainty quantification \cite{abdar_review_2021, hullermeier_aleatoric_2021, kendall_what_2017} are yet to be solved or even understood properly. Indeed, while neural networks may be highly accurate on benchmark and test datasets, no formal guarantees on their behaviour ``in the wild" can be provided to the end-user. 

These shortcomings stand in the way of wider scale adoption of deep neural networks in industrialized precision agriculture solutions. The typical user, who does not fully understand the models nor is provided with guarantees on them, has difficulty in trusting the systems \cite{condran_machine_2022, saia_transitioning_2020}.

To tackle one angle of this multi-dimensional problem, we propose to focus in this article on the problem of uncertainty quantification and control. Can we quantify the uncertainty of neural networks predictions? Can we provide valid guarantees on the performance of neural networks under real-world conditions so as to cultivate trust in systems that include these predictive models?

Conformal prediction \cite{vovk_algorithmic_2005, shafer_tutorial_2008} offers an interesting framework for producing predictions with valid statistical guarantees, and quantifying a black box model's uncertainty \cite{shi_applications_of_class_conditional, balasubramanian_conformal_2014}. In a context of classification, this framework allows a predictive model to produce prediction sets for a given observation $X$, instead of point predictions, with guarantees that the true value $Y$ is included in the prediction set with high probability. Concretely, given a specified error tolerance level $\alpha \in [0, 1]$, conformal prediction produces prediction sets $\mathcal{C}_{1-\alpha} \in \mathcal{Y}$ that satisfy the marginal coverage property
\begin{equation}
    \mathbb{P} \big( Y \in \mathcal{C}_{1-\alpha} (X) \big) \ge 1 - \alpha
    \label{eq1:marginal_coverage}
\end{equation}
For example, if the user sets $\alpha$ to $10\%$, then the conformal model will produce prediction sets that guarantee that the true value is predicted $90\%$ of the times.

Although useful and intuitive in its basic form, the original conformal approach only guarantees the results marginally; that is, on average over all observations. It does not provide any guarantees on specific subsets of observations: a property that would be quite useful in agricultural applications. Indeed, it is more important to provide performance guarantees for a given species or on the user's specific parcel and conditions rather than on average everywhere. 

For this reason, ``group-conditional conformal prediction" has been developed, with the aim of providing equalized coverage guarantees for all groups of individuals \cite{vovk_conditional_2013, romano_malice_2020}. Formally, let every individual be defined by the triplet $(X, Y, G) \in \mathcal{X} \times \mathcal{Y} \times \mathcal{G}$ where $G$ is the group, then group conditional conformal prediction aims at producing prediction sets $\mathcal{C}_{1-\alpha, g}$ with the following group-conditional coverage guarantee: 
\begin{equation}
    \mathbb{P} \big( Y \in \mathcal{C}_{1-\alpha, g}(X) | G = g \big) \ge 1 - \alpha \;\;\;\;\;\; \forall g \in \mathcal{G}
    \label{eq2:cond_coverage}
\end{equation}

The current article explores the application of group-conditional conformal prediction in an agricultural context; specifically, on the problem of crop and weed image classification using neural networks with the existence of auxiliary metadata describing various environmental and climatic characteristics of the image's content and context. The article's contributions can be summarized as being:
\begin{itemize}
    \item introduction and presentation of conformal prediction methods to the agricultural community concerned by uncertainty in machine learning-based decision-making;
    \item application of the marginal adaptive prediction sets (APS) \cite{romano_classification_2020, angelopoulos_uncertainty_2021} method to our classification use case, providing marginal coverage guarantees that will be shown empirically;
    \item simple description of the ``classical" group-conditional APS approach via iterative group-specific calibration and prediction \cite{romano_classification_2020, angelopoulos_conformal_2023};
    \item proposal of a simple and elegant alternative for a more efficient group-conditional calibration via quantile regression.
\end{itemize}

The article is structured as follows: Section \ref{sec:conf_pred} sets up the mathematical framework for the rest of the article then presents conformal prediction in its general form, with a focus on Adaptive Prediction Sets, a method that guarantees marginal coverage. Section \ref{sec:exper} presents the experimental setup and the results of marginal APS on the problem of image classification into weed and multiple crops on a large dataset. The results are presented in the light of environmental auxiliary variables thus showing the insufficiency of marginal coverage for the agricultural applications of interest. Section \ref{sec:group_cond} explores the group-conditional extension to conformal prediction and presents the group-conditional APS approach. A reformulation of the group quantile estimation procedure as quantile regression on group membership is then developed. The results of the classical iterative and quantile regression approaches are shown on a number of auxiliary variables chosen to form groups. Section \ref{sec:conclusion} concludes with a discussion of the results and a future vision of conformal prediction, particularly for agricultural applications.

\section{Conformal Prediction} \label{sec:conf_pred}
\subsection{Notation \& Setup} \label{sec:notation}
Before we dive into the details of the conformal approach, we define the mathematical setup that will be used in the rest of the article. 
We are in a supervised learning context, whereby for each input (image) $X \in \mathcal{X}$ is associated a ground-truth class $Y \in \{1,..., K \}$. As in a typical learning framework, we observe a sample of $N$ observations which we split into training and validation sets. Let $\mathcal{I}_1$ be the set of training observations. In the split-conformal framework \cite{papadopoulos_inductive_2002}, we further divide the validation set into two datasets; namely, the calibration set $\mathcal{I}_2$ and the proper test set $\mathcal{I}_3$. Note that $\mathcal{I}_1$, $\mathcal{I}_2$ and $\mathcal{I}_3$ are mutually exclusive.

We train on $\mathcal{I}_1$ a neural network classifier $\mathcal{B}$ that produces for each input $X \in \mathcal{X}$ a predicted output $\hat{y} \in \{1,..., K \}$. We also have access to the softmax output for each class at the last layer of the neural network; we call them $p^{[1]},..., p^{[K]}$.

\subsection{General Presentation} \label{sec:gen_pres}

First formulated by Vovk \textit{et al.} \cite{vovk_algorithmic_2005}, conformal prediction is an uncertainty quantification and control technique based on frequentist statistics. Broadly speaking, it can be understood as a method that allows the prediction of ``confidence intervals", instead of point predictions, at a specified level of significance $1-\alpha$. These prediction sets are guaranteed to contain the true value at least $1-\alpha$ of the times. This is the \textit{marginal coverage guarantee} presented in Equation \ref{eq1:marginal_coverage}. The only condition required for the validity of these methods is the \textit{exchangeability} of observations, which is a slightly weaker condition than the i.i.d. assumption commonly considered in statistical frameworks \cite{shafer_tutorial_2008}.

\begin{figure}[h]
\begin{center}
   \includegraphics[width=\linewidth]{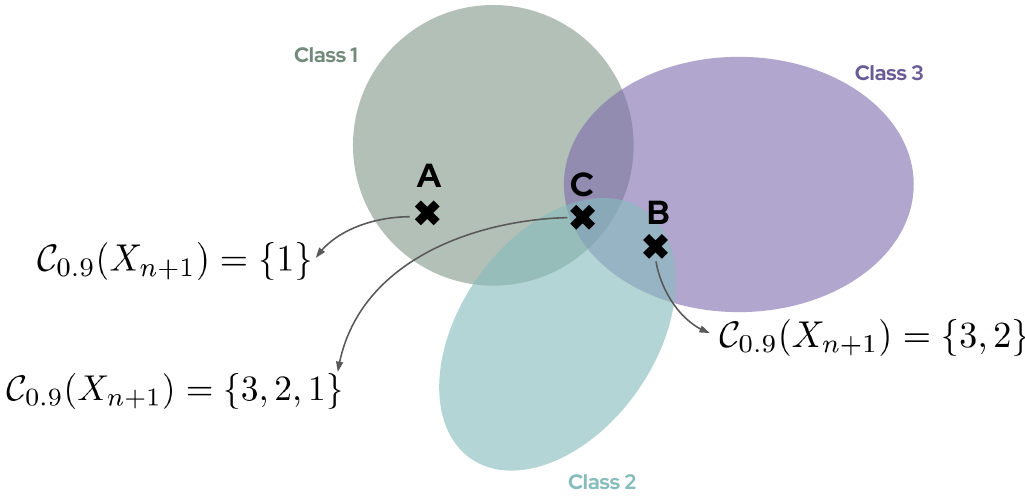}
\end{center}
   \caption{Representation of conformal prediction sets for three points, $A$, $B$ and $C$, with different levels of uncertainty.}
\label{fig:conf_represent1}
\end{figure}

For a new input $X_{n+1} \in \mathcal{I}_3$, a conformal algorithm compares this input, using a measure of conformity (that will be defined later), to the calibration set $\mathcal{I}_2$ of observations that the conformal model has previously seen. Based on the conformity of $X_{n+1}$, the conformal model will be more or less confident in its prediction, as such predicting a conformal set that is more or less large in such a way as to guarantee the existence of the true value inside.

Consider the representation space shown in Figure \ref{fig:conf_represent1} where we wish to predict conformal sets at the $90\%$ level of confidence. If a new input $X_{n+1}$ falls in position \textit{A}, it is clearly in the domain of Class 1. The conformal model can predict with high confidence only one class while guaranteeing a high coverage at $90\%$. If the new input appears in the more ambiguous region at position \textit{B}, then the conformal model, will produce a bigger prediction set with two classes in order to maintain the coverage guarantee at the desired $90\%$ level. Finally, if the new input is a difficult example and falls in the region with high uncertainty at point \textit{C}, then the model will predict all the classes in such a way as to guarantee predicting the true value. 

    \begin{figure*}[!t]
        \begin{center}
            \includegraphics[width=\textwidth]{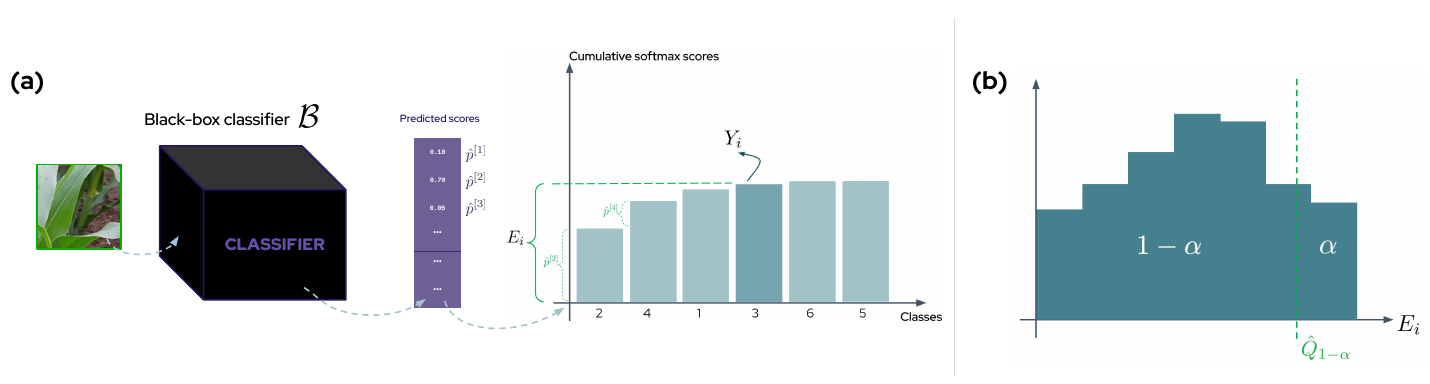}
            \caption{APS calibration: (a) Computation of the $E$ score for a given input; (b) Estimation of the decision quantile on the empirical distribution of scores.}
            \label{fig:aps_calibration}
        \end{center}
    \end{figure*}

\subsection{Adaptive Prediction Sets (APS)}
\label{sec:aps}

\begin{figure}[h]
        \begin{center}
           \includegraphics[width=\linewidth]{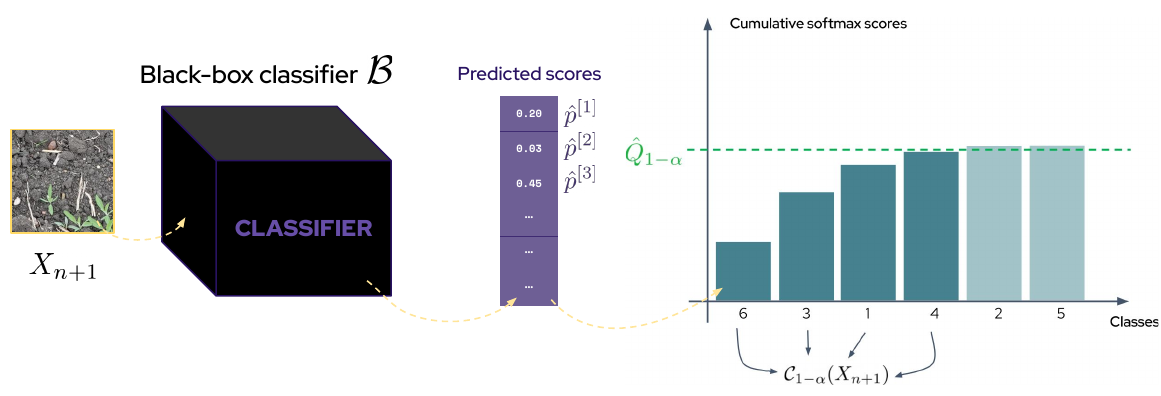}
        \end{center}
           \caption{APS prediction.}
        \label{fig:aps_prediction}
    \end{figure}

First proposed by \cite{romano_classification_2020} then improved and adapted to neural network classifiers in \cite{angelopoulos_uncertainty_2021}, the APS method not only provably achieves the marginal coverage guarantee but is also designed so that the size of the prediction sets \textit{adapts} to the ``difficulty" (think, \textit{uncertainty}) of each example. As such, it provides both a global measure of model uncertainty and also an individual-level measure of uncertainty where bigger predicted sets indicate higher model uncertainty. The approach follows the typical split-conformal procedure of calibration and prediction:

\begin{enumerate}
    \item \textsc{Calibration Step} \\
    After training the neural network on $x_j, j \in \mathcal{I}_1$, we now pass every individual $x_i, i \in \mathcal{I}_2$ into the network and compute its ``conformity score" defined as:
    \begin{equation}
        E_i = \sum_{t=1}^T p_i^{(t)}
    \end{equation}
    where the softmax scores are ordered in decreasing order, $t$ being the rank of the $t^{\text{th}}$ class with highest softmax output, and $T$ the rank of the ground-truth class. Accordingly, the conformity score is the cumulative softmax score of individual $i$ until reaching its true class $y_i \in \{1,..., K \}$. $E_i$ is thus the cumulative pseudo-probability mass assigned to the true class by the neural network (see Figure~\ref{fig:aps_calibration}(a)). In general, the bigger the probability mass, the more difficulty the neural network is having in finding the true class. For the specific case where the true class is predicted with a softmax score close to 1, see \cite{angelopoulos_uncertainty_2021} for a regularized version that allows such a class not to be rejected. 

    After obtaining the scores on the calibration set, we estimate $\hat{Q}_{1-\alpha}$, the $1-\alpha$ quantile of the empirical distribution of these scores, as can be seen in Figure~\ref{fig:aps_calibration}(b). This quantile is the maximum score among the $1 - \alpha$ lowest scores assigned to the true class by the neural network. It will be used to construct the prediction sets in the next step of the procedure.

    \item \textsc{Prediction Step} \\
    For the previously unseen inputs from the prediction set $\mathcal{I}_3$, we can now construct conformal sets by passing each individual in the neural network and comparing the score $p^{(k)}$ of each class $k$ to the estimated quantile $\hat{Q}_{1-\alpha}$. The classes that will form the prediction set are those classes whose cumulative softmax scores do not  exceed $\hat{Q}_{1-\alpha}$ (Figure \ref{fig:aps_prediction})\footnote{This is a slight simplification of the procedure, refer to \cite{angelopoulos_uncertainty_2021, romano_classification_2020} on the importance of randomizing the inclusion of the classes around the decision quantile.}. These are the classes that are considered ``probable" enough to be predicted. The notion of ``conformity score" discussed in Section \ref{sec:gen_pres} appears here: indeed, scores that are higher than $\hat{Q}_{1-\alpha}$ are considered \textit{non-conformal} (think, ``too extreme", or ``too improbable"), and as such are not considered to be valid predictions.

\end{enumerate}

\section{Experimental Setup} \label{sec:exper}
\subsection{Data}
\begin{figure}[h]
        \begin{center}
           \includegraphics[width=\linewidth]{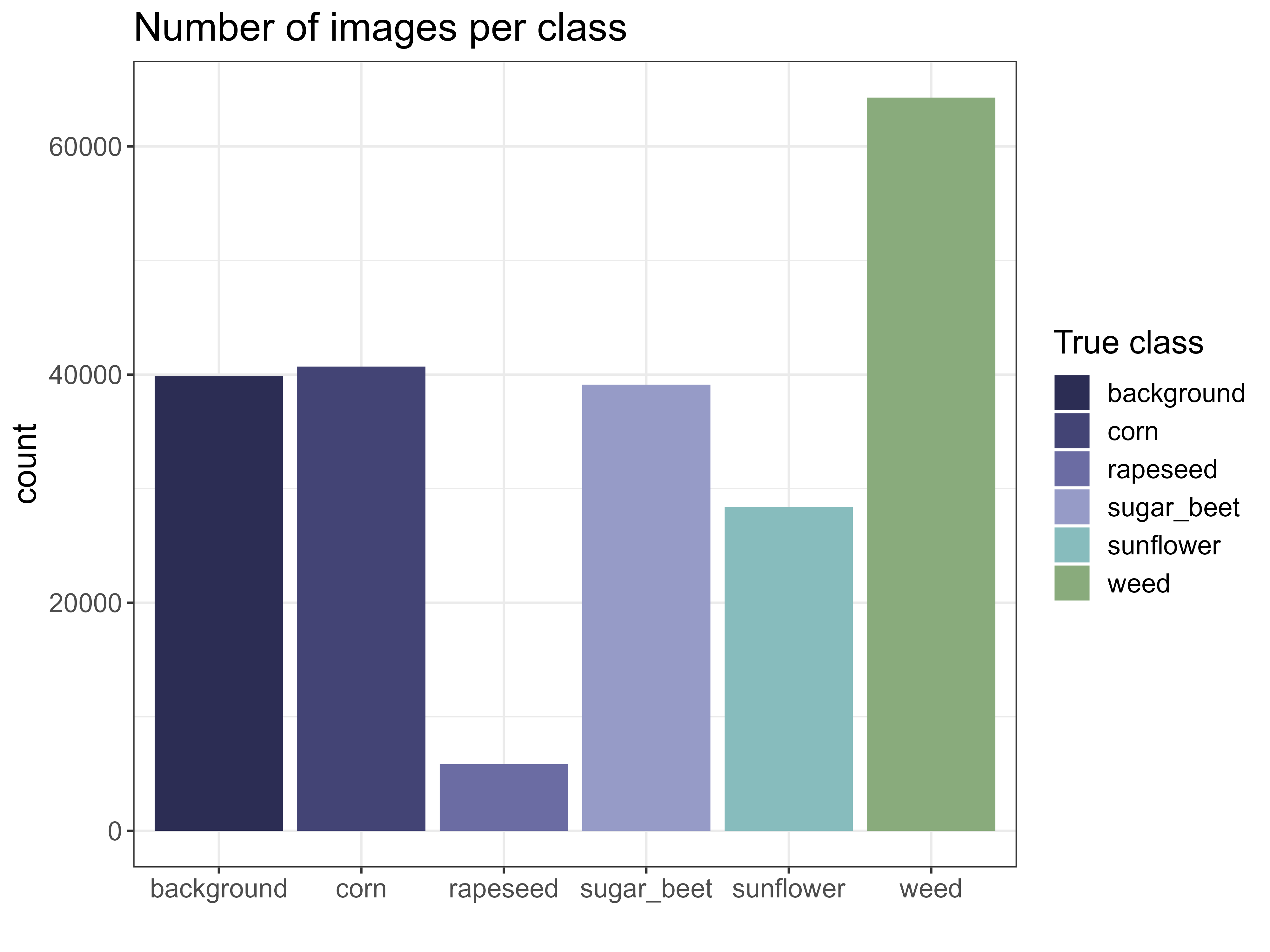}
        \end{center}
           \caption{Distribution of images over the 6 classes.}
        \label{fig:classes_barplot}
    \end{figure}

To demonstrate the conformal approaches on an agricultural use case, we work on a specialized proprietary dataset gathered in multiple locations around the world, under real-world uncontrolled conditions, for the problem of visual identification of crop and weed via image classification. The dataset consists of 218 thousands RGB images of size $224 \times 224$ annotated internally. Associated to each image is one of six classes specifying the crop type of the largest ``object" in the image: \textit{corn}, \textit{rapeseed}, \textit{sugar beet}, \textit{sunflower} or \textit{weed}. A final class \textit{background} is assigned to the images where there is no plant. The distribution of the images over the different classes can be seen in Figure \ref{fig:classes_barplot}.

\subsection{Auxiliary Data}
\label{sec:aux_data}
To each image is associated a number of auxiliary variables (``metadata") that describe different factors related to the image. Some of these factors can be considered ``intrinsic" to the visual scene -- that is, visible -- such as some pedoclimatic characteristics like the color, texture and humidity of the soil. Other variables describe the broader environmental characteristics that may have direct or indirect influence on the image such as the conditions of the sky and the wind or the geographical location of the acquisition. These metadata are entered at the moment of image capture by the data acquirers based on their qualitative evaluation of the conditions following well-defined criteria. The visually-verifiable metadata are also reviewed during the annotation process. Other metadata such as geo-location, time and sensor conditions are automatically captured and saved.

For the purpose of the current article, with the aim of keeping the presentation as concise and clear as possible, we focus on only two auxiliary variables that are particularly interesting for practical use cases:
\begin{itemize}
    \item \textit{Location:} it is the location of the acquisitions as defined using GPS coordinates. From a broad perspective, our data can be divided into eight different locations across Europe denoted A to H. High-level positioning of these locations can be seen in Figure \ref{fig:locations}. Given that each location is characterized by largely different environmental and pedo-climatic conditions, this auxiliary variable can be considered a proxy for multiple other characteristics and holds high practical interest: it is important to guarantee acceptable levels of detection in all locations where the system is to be deployed.
    \begin{figure}
        \begin{center}
           \includegraphics[width=\linewidth]{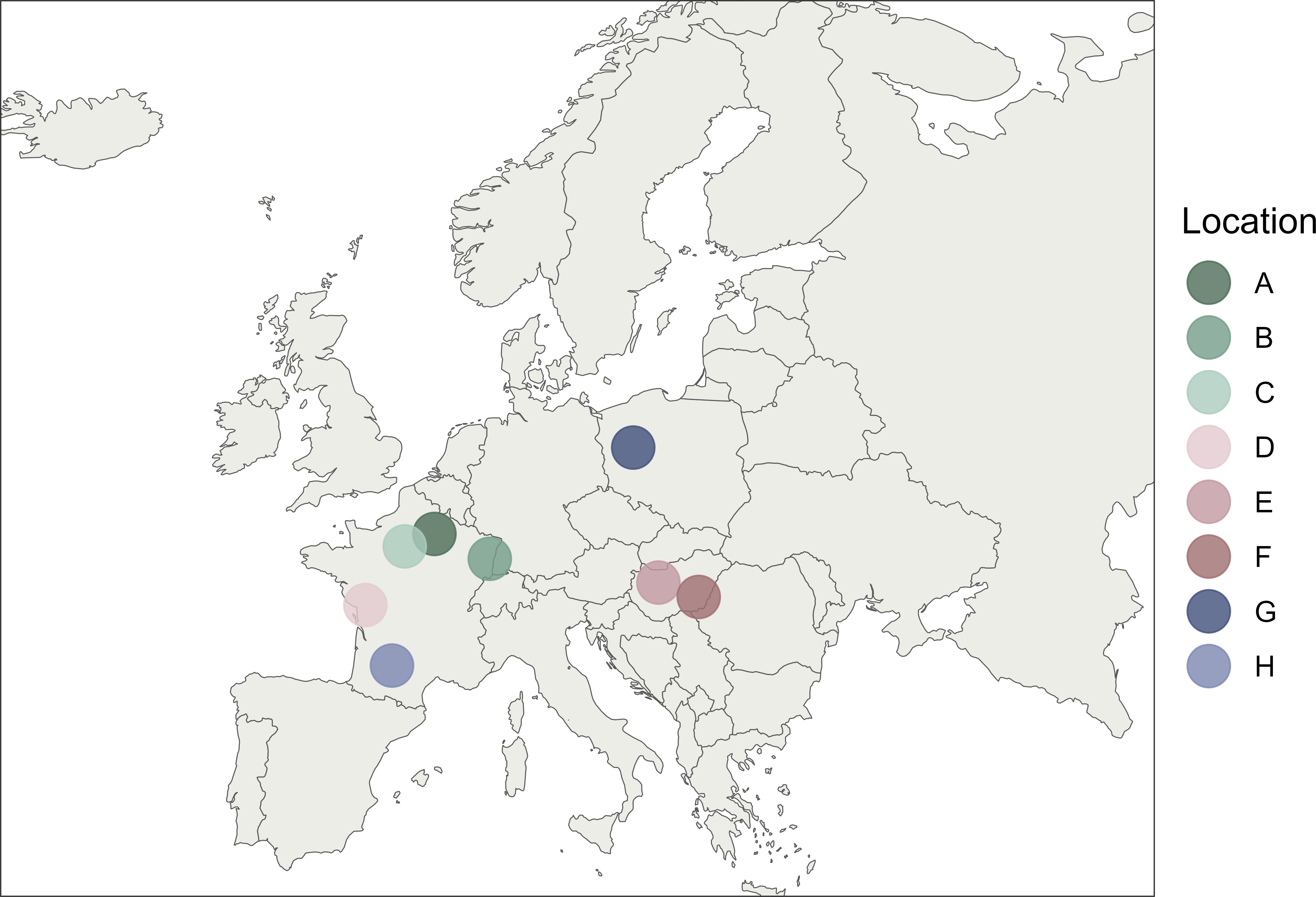}
        \end{center}
           \caption{Locations of data acquisition in Europe.}
        \label{fig:locations}
    \end{figure}

    \item \textit{Sky:} this variable represents the ``perceived" condition of the sky at the moment of data acquisition. It can take one of two values, of each an example is shown in Figure \ref{fig:sky_conditions}: \texttt{overcast} (a) and \texttt{sunny} (b). The condition of the sky has an interesting impact on the visual characteristics of the image, such as luminosity, color temperature and shadows. Since the system is to be deployed in uncontrolled environments, detection results should be guaranteed regardless of the sky and ambient light.

    \begin{figure}
        \begin{center}
           \includegraphics[width=0.65\linewidth]{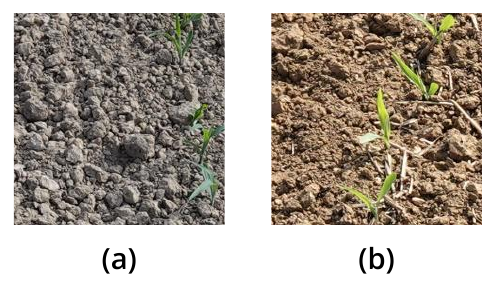}
        \end{center}
           \caption{Examples of images taken in different \textit{sky} conditions in the same \textit{location}: (a) \texttt{overcast}, (b) \texttt{sunny}.}
        \label{fig:sky_conditions}
    \end{figure}
\end{itemize}

\subsection{Base Model}
As mentioned previously, conformal prediction requires a base predictor that produces point predictions, which will be ``transformed" via the conformal procedure into a conformal predictor producing sets of prediction points. For the purpose of this study and without loss of generality, the base classifier used is a classic ResNet18 network \cite{he_deep_2016} pre-trained on ImageNet \cite{deng2009imagenet} and fine-tuned on our training data. It is important to note that the proposed conformal approaches are independent of the chosen base classifier. It can be any neural network architecture or other model such as random forests or support vector machines \cite{romano_classification_2020}.

\subsection{Experimental Results: Marginal APS}
\label{sec:aps_exp_results}
\begin{table}[h!]
\begin{center}
\begin{tabular}{|l|c|c|}
\hline
Method & Coverage & Set Size \\
\hline\hline
Base (Top-1 Accuracy) & 0.680 & 1.000 \\
Marginal APS Classifier &  0.896  & 2.566 \\
\hline
\end{tabular}
\end{center}
\caption{Comparison of Base \& Conformal ResNet18 classifier.}
\label{tab:base_vs_conformal}
\end{table}

We finetune the ResNet18 network on the training set  $\mathcal{I}_1$ (50\% of the database), then calibrate and predict respectively on $\mathcal{I}_2$ and $\mathcal{I}_3$ (45\% and 55\% of the remaining individuals) respectively following the APS procedure with an error tolerance level fixed at $\alpha = 0.1$.

Table \ref{tab:base_vs_conformal} shows a comparison between the coverage obtained for the base classifier with its conformal version. The coverage of the base point predictor (which corresponds to its Top-1 overall accuracy) is 68\%, with a unique set size of 1, since we only predict the top class. The APS procedure maintains the coverage exactly at the required $1-\alpha = 0.9$ level, with an average prediction set size of 2.6. That is, by calibrating the predictive system on a dataset that resembles the population on which we want to predict and  permitting the network to predict, on average, between 2 and 3 classes, we guarantee finding the true class 90\% of the time. 

\begin{table}[ht]
\begin{center}
\begin{tabular}{|c|c|c|c|c|}
  \hline
 Group & \textit{Location} & \textit{Sky} & Coverage & Set Size \\ 
  \hline
  1 & \texttt{A} & \texttt{overcast} & 0.935 & 2.823 \\ 
  2 & \texttt{A} & \texttt{sunny} & 0.872 & 2.614 \\ 
  3 & \texttt{B} & \texttt{overcast} & 0.926 & 1.560 \\ 
  4 & \texttt{B} & \texttt{sunny} & 0.914 & 1.953 \\ 
  5 & \texttt{C} & \texttt{overcast} & 0.966 & 2.625 \\ 
  6 & \texttt{C} & \texttt{sunny} & 0.877 & 2.412 \\ 
  7 & \texttt{D} & \texttt{overcast} & 0.891 & 2.476 \\ 
  8 & \texttt{D} & \texttt{sunny} & 0.901 & 2.437 \\ 
  9 & \texttt{E} & \texttt{overcast} & 0.944 & 2.105 \\ 
  10 & \texttt{E} & \texttt{sunny} & 0.908 & 2.450 \\ 
  11 & \texttt{F} & \texttt{overcast} & 0.959 & 2.454 \\ 
  12 & \texttt{F} & \texttt{sunny} & 0.937 & 2.741 \\ 
  13 & \texttt{G} & \texttt{overcast} & 0.990 & 2.728 \\ 
  14 & \texttt{G} & \texttt{sunny} & 0.943 & 2.348 \\ 
  15 & \texttt{H} & \texttt{sunny} & 0.943 & 2.477 \\ 
  \hline
  \multicolumn{3}{|c|}{Marginal APS Classifier} &  \textbf{0.896}  & \textbf{2.566} \\

   \hline
\end{tabular}
\end{center}
\caption{Results of the Marginal APS classifier, per group.}
\label{tab:aps_per_group}
\end{table}

Although the coverage is perfectly maintained marginally, the picture changes when we look at the conditional coverage per group. What if we like to guarantee the $1-\alpha$ coverage for each possible agro-environmental condition in our data? 

As Table \ref{tab:aps_per_group} shows, the coverage is not maintained at the desired level but is highly varying among the groups (Note that the group \texttt{H,overcast} is not included in the table because this combination does not exist in the data). Although the group-conditional coverage criterion defined in Equation \ref{eq2:cond_coverage} does not seem, empirically, to be violated for a number of groups, we cannot say that the condition is guaranteed since there are no explicit constraints on the estimation of the quantile or the construction of the prediction sets in such a way as to provide such a guarantee. Indeed, for such groups as Group 2 and Group 6, the coverage is far from being maintained; while for other groups we see that the coverage is overly conservative leading to bigger prediction sets (in size) that may be required.

\section{Group-conditional Conformal Prediction} \label{sec:group_cond}
The marginal coverage guarantee may not be useful in a number of use cases since it does not imply validity on all individuals; that is, conditional on their idiosyncratic characteristics. While the coverage is maintained on average, it is not guaranteed on certain groups of individuals; usually those that are not represented enough in the data \cite{romano_malice_2020}. In a number of use cases, such as the deployment of an autonomous weed detection system in new environments or the detection of diseases in plants, it is required to provide guarantees on all groups of individuals so that the system may be deemed reliable. Group-conditional conformal prediction has been developed for this purpose, providing the conditional coverage guarantee defined in Equation \ref{eq2:cond_coverage}. 

Now that we have defined the notion of auxiliary variables in Section \ref{sec:aux_data}, we can refine the definition of a ``group." Assume that for each individual we observe an image $X$ to which we associate a ground-truth label $Y$, and a number of auxiliary variables $\{M_L \in \mathcal{M}_L, M_S \in \mathcal{M}_S, ... \}$. An individual's group is thus defined as being its observed combination of auxiliary data: $G \in \mathcal{G}$, where $\mathcal{G} = \mathcal{M}_L \times \mathcal{M}_S \times ...$ . For the sake of simplicity and without loss of generality, we assume that we only observe the two auxiliary variables \textit{location} and \textit{sky}. For example, one group can be defined as $G_1 = \{M_L = \text{\texttt{A}}, M_S = \text{\texttt{overcast}} \}$. We can thus provide the coverage guarantee: 
\begin{equation}
\begin{split}
    \mathbb{P} \big( Y \in \mathcal{C}_{1-\alpha, G_1}(X) &| M_L = \text{\texttt{A}}, M_S = \text{\texttt{overcast}} \big) \\ 
    & \ge 1 - \alpha
    \label{eq2:ex_cond_coverage}
\end{split}
\end{equation}

\subsection{Iterative Group-conditional APS}
The ``classical" approach to produce prediction sets that satisfy the group coverage guarantee consists of iteratively conducting the APS Calibration procedure described in Section \ref{sec:aps} and Figure \ref{fig:aps_calibration} on each group $g \in \mathcal{G}$ separately \cite{angelopoulos_conformal_2023}. A conformal decision quantile $\hat{Q}_{1-\alpha}^{(g)}$ is estimated separately for each group $g$ on the individuals in $\mathcal{I}_2$ that satisfy the conditions of group $g$.

Then, for a new individual whose auxiliary variables are observed, we simply produce a prediction set following the APS Prediction procedure using the group-specific $\hat{Q}_{1-\alpha}^{(g)}$ quantile. Although quite simple to implement and understand, such a method may prove to be time inefficient, especially for a large number of groups, since it requires an iterative traversing and quantile estimation on each group separately.

\subsection{Calibration by Quantile Regression}
\label{sec:quant_reg}
We propose a simple and more elegant reformulation of the group-conditional conformal calibration procedure via quantile regression. Quantile regression \cite{koenker_regression_1978, beyerlein_quantile_2014} is a method that allows the estimation of a desired $\tau \in [0, 1]$ quantile of a dependent variable $Y$ based on a set of explanatory variables $X$ \footnote{Note that $X$ and $Y$ here are not as defined previously but are generic variable names in keeping with common definitions of regression models.}. It can be understood as the counterpart of linear regression -- that estimates the mean of the output variable -- for the estimation of the quantiles, a special case of which is the median for $\tau = 0.5$. For an output variable $Y$ and explanatory variables $X$, a generic formulation of the quantile regression is given by:
\begin{equation}
    Q_{Y|X}(\tau) = X\beta_\tau
\end{equation}
where $Q_{Y|X}(\tau)$ is the $\tau$ quantile of the conditional distribution of $Y$ given $X$, assuming a linear relationship between the conditional quantile and the explanatory variables. The estimated coefficient $\hat{\beta}_\tau$ is solution to the following optimization problem: 
\begin{equation}
\begin{split}
    \hat{\beta}_{\tau} =  \argmin_{\beta \in \mathbb{R}^d} \big[ &(\tau - 1) \sum_{Y_i < X_i\beta} (Y_i - X_i \beta) \; + \\
                                        & \phantom{(1-}\tau \;\;\; \sum_{Y_i > X_i\beta} (Y_i - X_i \beta) \big]
\end{split}
\end{equation}
where $d$ is the dimension of the vector $X$. This minimization problem can be efficiently solved using linear programming approaches \cite{koenker_algorithm_1987, koenker2009quantreg}.

\subsubsection{Calibration}
    We can thus estimate the group-conditional $1-\alpha$ quantiles of the scores by regressing them on group-membership indicator variables. This constitutes the calibration of the conformal procedure.

\begin{figure*}[!h]
        \begin{center}
            \includegraphics[width=\textwidth]{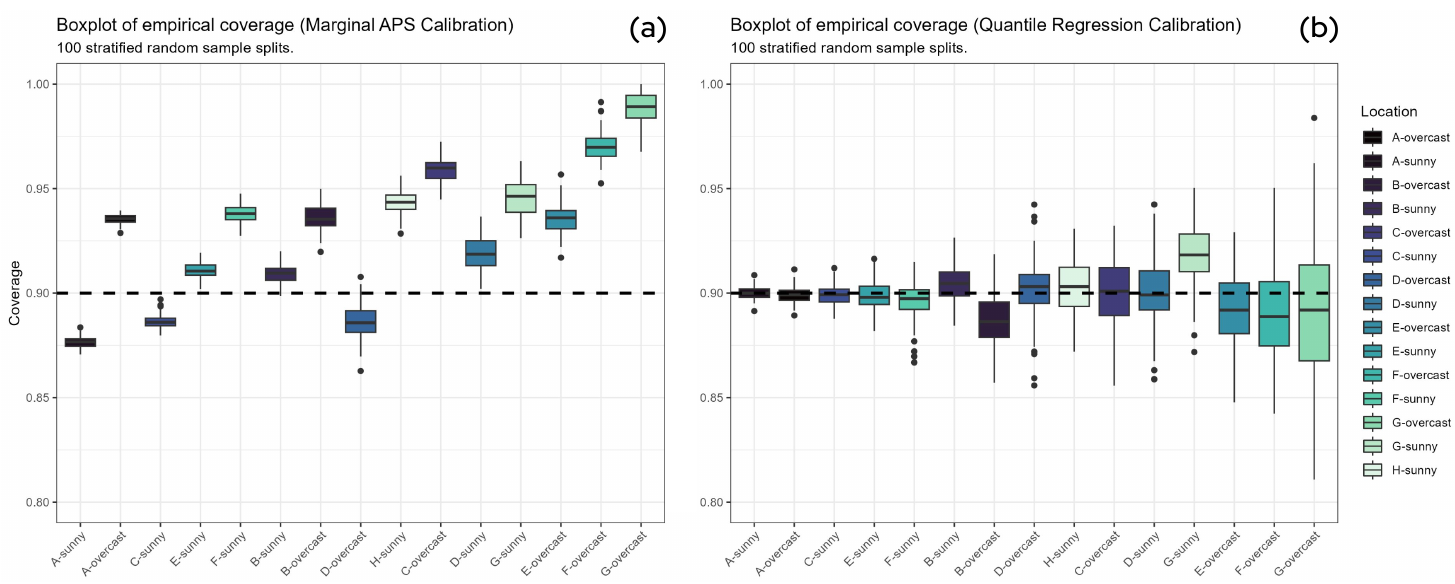}
            \caption{Boxplot of the empirical coverage per group over 100 different splits of the validation set: (a) Marginal APS. (b) Quantile Regression Calibration (ours). Groups are sorted in decreasing order of number of individuals.}
            \label{fig:qreg_vs_marginal}
        \end{center}
    \end{figure*}

To illustrate how the approach works, we consider the two previously described auxiliary variables, \textit{location} and \textit{sky}. To simplify the presentation, we consider that the variable \textit{location} has only two levels: $M_L \in \{\texttt{A}, \texttt{B} \}$, and \textit{sky}: $M_S \in \{\texttt{sunny}, \texttt{overcast} \}$. The regression model
\small
\begin{equation}
    \begin{split}
        Q_{E | \{M_L, M_S \}}(1 - \alpha) = & \; \beta_0 \; + \\
        & \; \beta_{\texttt{A}} \mathds{1}_{\{M_L = \texttt{A} \}} \; + \\
        & \; \beta_{\texttt{sunny}} \mathds{1}_{\{M_S = \texttt{sunny} \}} \; + \\
        & \; \beta_{\texttt{A,sunny}} \mathds{1}_{\{M_L = \texttt{A} \}} \mathds{1}_{\{M_{S} = \texttt{sunny} \}}
    \end{split}
\end{equation}
\normalsize
where $\mathds{1}_{\{ \; \}}$ is the indicator function, thus allows us to estimate the $1-\alpha$ quantile of the scores for all the possible groups defined by these two auxiliary variables. Notice that all the groups are identified in this model: $\hat{\beta}_0$, the estimated intercept, is the estimated $1-\alpha$ quantile of the score for the baseline group, defined by the conditions that are not explicitly specified in the regression equation; in this case $\{M_L = \texttt{B}, M_{S} = \texttt{overcast} \}$. The other estimated coefficients $\hat{\beta}_\texttt{A}$, $\hat{\beta}_{\texttt{sunny}}$ and $\hat{\beta}_{\texttt{A,sunny}}$ are to be interpreted as the difference in quantiles from the baseline $\hat{\beta}_0$. Hence, the estimated quantile for the group $\{M_L = \texttt{B}, M_{S} = \texttt{sunny} \}$ is $\hat{\beta}_0 + \hat{\beta}_{\texttt{sunny}}$, just as the estimated quantile of the group $\{M_L = \texttt{A}, M_{S} = \texttt{sunny} \}$ is $\hat{\beta}_0 + \hat{\beta}_\texttt{A} + \hat{\beta}_{\texttt{sunny}} + \hat{\beta}_{\texttt{A,sunny}}$.

This methodology can be simply expanded for the case where more auxiliary variables are considered or where the auxiliary variables have more than two levels, or are continuous \cite{koenker_regression_1978} -- unlike the classical approach.

\subsubsection{Prediction}

For a new observation for which we observe the auxiliary data, we can easily plug-in its values in the regression model and obtain its corresponding quantile estimation. It is the estimated quantile of the group to which the observation belongs. The obtained $\hat{Q}_{1-\alpha}^{(g)}$ will then be used following the APS prediction procedure previously described in Section \ref{sec:aps} and Figure \ref{fig:aps_prediction} to produce prediction sets for this new observation.

\subsubsection{Experimental Results}
The proposed approach is compared to the Marginal APS. The validation set is split into a calibration set $\mathcal{I}_2$ (45\%) and prediction set $\mathcal{I}_3$ (55\%) following a stratified proportion sampling scheme where each group is sampled according to its proportion in the validation set. The two methods are calibrated and tested on the same data. In order to validate the results, we implement a resampling scheme over 100 iterations leading to a different split of the validation set at each iteration.

Figure \ref{fig:qreg_vs_marginal} shows the boxplots of the obtained coverage per group for the 100 resamplings for the two methods, with the groups sorted by decreasing order of number of individuals. While the Marginal APS shows, generally, a smaller variance per group, its group-specific coverages are highly biased. We observe a high variability in the group coverages, echoing the results previously presented in Section \ref{sec:aps_exp_results}. On the other hand, our proposed group-conditional method stably maintains the group coverage at the required 0.9 level, on average, for all groups. Even though the variance of the observed coverage is naturally higher for less-represented groups, it is still acceptably maintained over the 100 iterations.

Table \ref{tab:qreg_vs_aps} shows the average empirical coverage and set size for each group over the 100 resamplings. The proposed approach by quantile regression leads, on average, to smaller prediction sets without compromising on coverage. The importance of such a result may not be obvious in use cases with few classes like the current one. However, on datasets with a large number of classes, valid prediction sets with smaller size are largely preferred for use cases of automated decision making based on the predicted sets, and applications requiring a study of the prediction sets by a human agent \cite{angelopoulos_uncertainty_2021}.

\begin{table*}[!ht]
\begin{center}
\begin{tabular}{|c|c|c|c|c|c|c|}
  \hline
  \multicolumn{3}{|c|}{} & \multicolumn{2}{|c|}{\textbf{Marginal APS}} & \multicolumn{2}{|c|}{\textbf{Quantile Regression}} \\
  \hline
 Group & \textit{Location} & \textit{Sky} & Coverage & Set Size & Coverage & Set Size \\ 
  \hline
  1 & \texttt{A} & \texttt{overcast} & 0.935 & 2.838 & 0.899 & 2.415 \\  
  2 & \texttt{A} & \texttt{sunny} & 0.876 & 2.627 & 0.900 & 2.893 \\  
  3 & \texttt{B} & \texttt{overcast} & 0.936 & 1.566 & 0.887 & 1.354 \\  
  4 & \texttt{B} & \texttt{sunny} & 0.909 & 1.985 & 0.904 & 1.939 \\  
  5 & \texttt{C} & \texttt{overcast} & 0.959 & 2.615 & 0.900 & 1.929 \\  
  6 & \texttt{C} & \texttt{sunny} & 0.886 & 2.431 & 0.899 & 2.560 \\  
  7 & \texttt{D} & \texttt{overcast} & 0.886 & 2.482 & 0.901 & 2.683 \\  
  8 & \texttt{D} & \texttt{sunny} & 0.918 & 2.439 & 0.901 & 2.286 \\  
  9 & \texttt{E} & \texttt{overcast} & 0.936 & 2.157 & 0.892 & 1.735 \\  
  10 & \texttt{E} & \texttt{sunny} & 0.911 & 2.480 & 0.899 & 2.364 \\  
  11 & \texttt{F} & \texttt{overcast} & 0.970 & 2.494 & 0.892 & 1.699 \\  
  12 & \texttt{F} & \texttt{sunny} & 0.938 & 2.788 & 0.897 & 2.295 \\  
  13 & \texttt{G} & \texttt{overcast} & 0.989 & 2.707 & 0.891 & 1.682 \\  
  14 & \texttt{G} & \texttt{sunny} & 0.945 & 2.358 & 0.918 & 2.150 \\  
  15 & \texttt{H} & \texttt{sunny} & 0.943 & 2.493 & 0.903 & 2.066 \\  
  \hline
  \multicolumn{3}{|c|}{Marginal Results} &  \textbf{0.900}  & \textbf{2.570} & \textbf{0.898} & \textbf{2.137} \\

   \hline
\end{tabular}
\end{center}
\caption{Comparison of average empirical coverage and prediction set size over 100 different splits.}
\label{tab:qreg_vs_aps}
\end{table*}

\section{Conclusion} \label{sec:conclusion}
In this article, we introduced and presented the conformal prediction framework from a practical perspective with a special focus on its importance to the agricultural community. Indeed, as deep learning black box methods become the go-to approaches in a large spectrum of automated agricultural tasks, methods that provide valid guarantees on their performance -- or, at least, quantify the uncertainty associated to their predictions -- are important to certify their quality. Here, the work was demonstrated on the task of weed and crop classification in real-world conditions. Special attention has been accorded to the recently developed Adaptive Prediction Sets (APS) method which was shown to empirically maintain the marginal coverage guarantee as defined in Equation \ref{eq1:marginal_coverage}. However, the marginal guarantee is not enough to ensure the required coverage is maintained on all possible individuals or groups of individuals (in our case defined by auxiliary data acquired during image acquisition): it is thus not enough for multiple agricultural use cases. 

This motivated our presentation of group-conditional conformal prediction; first, via the classical approach that consists of iteratively applying the APS procedure on each group separately; then using our proposed ``elegant" approach via quantile regression of calibrated softmax scores on group membership indicators. The proposed approach allows for the joint estimation of the $1 - \alpha$ decision quantiles of all groups. Quantile regression calibration has been shown empirically to maintain the $1-\alpha$ coverage level for all groups, even those that are not largely represented in the dataset. This approach also provided smaller prediction sets, on average, per group being thus more useful from a decisional perspective -- simply because it is easier to take a decision when fewer classes are predicted.

This article is the first work, to the authors' knowledge, to introduce these notions and methods to the agri-tech community. It constitutes a first step in a research direction aiming at developing reliable and trustworthy machine learning systems on which the farmers can rely and have confidence in, even without fully understanding all their intricacies. Future work aims at extending the current methods to the more realistic scenario in which the auxiliary data are not, or only partially, observed on prediction images; at developing theoretical guarantees of the maintenance of group coverage by quantile regression; and finally at adapting and presenting the conformal methodology on other computer vision tasks such as object detection and image segmentation.

{\small
\bibliographystyle{ieee_fullname}
\bibliography{full_references}
}

\end{document}